\documentclass{article} 
\usepackage{colm2024_conference}

\usepackage{microtype}
\usepackage{amsmath}
\usepackage{hyperref}
\usepackage{url}
\usepackage{wrapfig}
\usepackage{graphicx}
\usepackage{booktabs}
\definecolor{darkblue}{rgb}{0, 0, 0.5}
\hypersetup{colorlinks=true, citecolor=darkblue, linkcolor=darkblue, urlcolor=darkblue}

\title{Conan-embedding: General Text Embedding with More and Better Negative Samples}

\author{
    Shiyu Li \textsuperscript{1,2}\thanks{This work was done when Shiyu Li was an intern at Tencent Platform and Content Group.} \qquad Yang Tang\textsuperscript{2} \qquad Shi-Zhe Chen\textsuperscript{2} \qquad Xi Chen\textsuperscript{2} \\
    \textsuperscript{1}ECE, Peking University \qquad
    \textsuperscript{2}BAC, Tencent PCG
    \\
    {\tt\small shiyuli@stu.pku.edu.cn} \qquad
    {\tt\small \{ethanntang, shizhechen, jasonxchen\}@tencent.com}
}

\colmfinalcopy 
\begin{document}

\maketitle

\begin{abstract}
With the growing popularity of RAG, the capabilities of embedding models are gaining increasing attention. 
Embedding models are primarily trained through contrastive learning, with negative examples being a key component. 
Previous work has proposed various hard negative mining strategies, but these strategies are typically employed as preprocessing steps. 
In this paper, we propose the conan-embedding model, which maximizes the utilization of more and higher-quality negative examples. 
Specifically, since the model's ability to handle preprocessed negative examples evolves during training, we propose dynamic hard negative mining method to expose the model to more challenging negative examples throughout the training process.
Secondly, contrastive learning requires as many negative examples as possible but is limited by GPU memory constraints.
Therefore, we use a Cross-GPU balancing Loss to provide more negative examples for embedding training and balance the batch size across multiple tasks.
Moreover, we also discovered that the prompt-response pairs from LLMs can be used for embedding training. Our approach effectively enhances the capabilities of embedding models, currently ranking first on the Chinese leaderboard of Massive text embedding benchmark \href{https://huggingface.co/spaces/mteb/leaderboard}{(MTEB)}.
\end{abstract}

\section{Introduction}

With the rapid development of natural language processing technology, embedding models~\cite{su2022one,bge_embedding,wang2023improving} have played a crucial role in text representation, information retrieval, and generation tasks. Embedding models map words, sentences, or documents into a high-dimensional continuous space, allowing similar texts to have closer vector representations. This representation not only enhances the operability of text data but also significantly improves performance in various downstream tasks. Particularly in retrieval-augmented generation (RAG) technology, the capability of embedding models directly affects the quality of generated results.

Despite significant progress in embedding models, existing methods still have shortcomings in negative example selection. Typically, embedding models are trained through contrastive learning, and the quality of negative examples is crucial to model performance. Previous research~\cite{Wang2022,moreira2024nv} has proposed various hard negative mining strategies, which have improved model performance to some extent. However, these strategies are mostly employed as preprocessing steps, limiting the model's performance when dealing with complex and variable training data.

To address these issues, this paper proposes the Conan-Embedding Model, maximizes the utilization of more and higher-quality negative examples. Specifically, we iteratively mine hard negatives during training, allowing the model to dynamically adapt to changing training data. Additionally, we introduce a cross-GPU balancing Loss to balance the number of negative examples across multiple tasks, improving training efficiency and effectiveness. We also discovered that the prompt-response pairs from large language models (LLMs) can be used as training data, further enhancing the performance of embedding models. With these improvements, our approach has achieved the first place on the Chinese Massive text embedding benchmark (CMTEB) leaderboard, demonstrating its outstanding performance and broad application prospects.

\section{Methods}
\begin{figure*}[h]
    \centering
    \includegraphics[width=\linewidth]{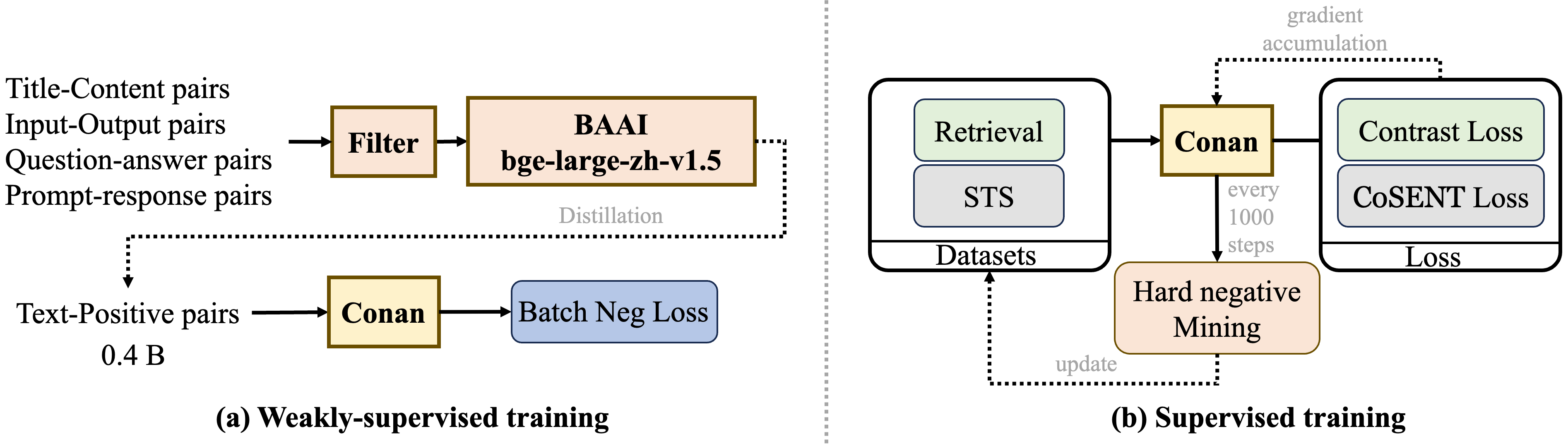}
    \caption{The pipeline of our methods includes both weakly-supervised and supervised training. During weakly-supervised training, we collect 0.75 billion pairs of datasets and select 0.4 billion of them. During supervised training, we use a dynamic hard negative mining strategy to better fine-tune the model.}
    \label{fig:framework} 
\end{figure*}

\subsection{Training Workflow}
\subsubsection{Pre-training}
Following~\cite{li2023towards}, we also use a multi-stage training approach. We divide the training into pre-training and fine-tuning stages. As shown in Figure.~\ref{fig:framework} (a), during the pre-training phase, we use standard data filtering methods as described in ~\cite{cai2024internlm2}. After filtering, we use the bge-large-zh-v1.5~\cite{bge_embedding} model for scoring, then we discard all data with scores below 0.4. To efficiently and fully utilize the pretrain data, we use InfoNCE loss with In-Batch Negative for training:
\begin{equation}
\mathcal{L}_{neg} = - \sum_{i=1}^N \log \frac{\exp(\text{sim}(x_i, y_i^+))}{\sum_{j=1}^M \exp(\text{sim}(x_i, y_i))}
\end{equation}
$x_i$ represents the query of the positive sample, $y_i^+$ represents the passage of the positive sample, and $y_i$ represents the passages of other samples in the same batch, which are considered as negative samples.

In-Batch Negative InfoNCE Loss~\cite{gutmann2010noise} is a loss function used for contrastive learning that leverages other samples within a mini-batch as negative examples to optimize the model. Specifically, in each mini-batch, all samples other than the target sample's positive pair are considered negative samples. By maximizing the similarity of the positive pairs and minimizing the similarity of the negative pairs, the In-Batch Negative InfoNCE Loss can effectively enhance the model's discriminative ability and representation learning performance. This method improves training efficiency and reduces the need for generating additional negative samples by fully utilizing the samples within the mini-batch.
\subsubsection{Supervised Fine-tuning}
At the supervised fine-tuning stage, we perform task-specific fine-tuning for different downstream tasks. As shown in Figure.~\ref{fig:framework} (b), we divide the tasks into two categories: retrieval and STS (semantic textual similarity). The retrieval task includes query, positive text, and negative text, with the classic loss function being InfoNCE loss. STS task involves distinguishing the similarity between two texts, with the classic loss function being cross-entropy loss. According to~\cite{kexuefm-9341} and other works~\cite{Moka2023}, CoSENT loss is slightly better than cross-entropy loss. Therefore, we also adopt CoSENT loss to optimize STS task, which is formulated as follows:
\begin{equation}
\mathcal{L}_{cos} = \log \left(1 + \sum\limits_{\text{sim}(i,j) > \text{sim}(k,l)} \exp \left( \frac{\cos(x_k, x_l) - \cos(x_i, x_j)}{\tau}\right) \right)
\end{equation}
where $\tau$ is the scale temperature, $\cos(\cdot)$ is the cosine similarity function, and
$sim(k, l)$ is the similarity between $x_i$ and $x_j$.

\subsection{Dynamic Hard Negative Mining}
\begin{wrapfigure}{h}{0.5\textwidth}
\vspace{-5pt}
\centering
\includegraphics[width=7cm]{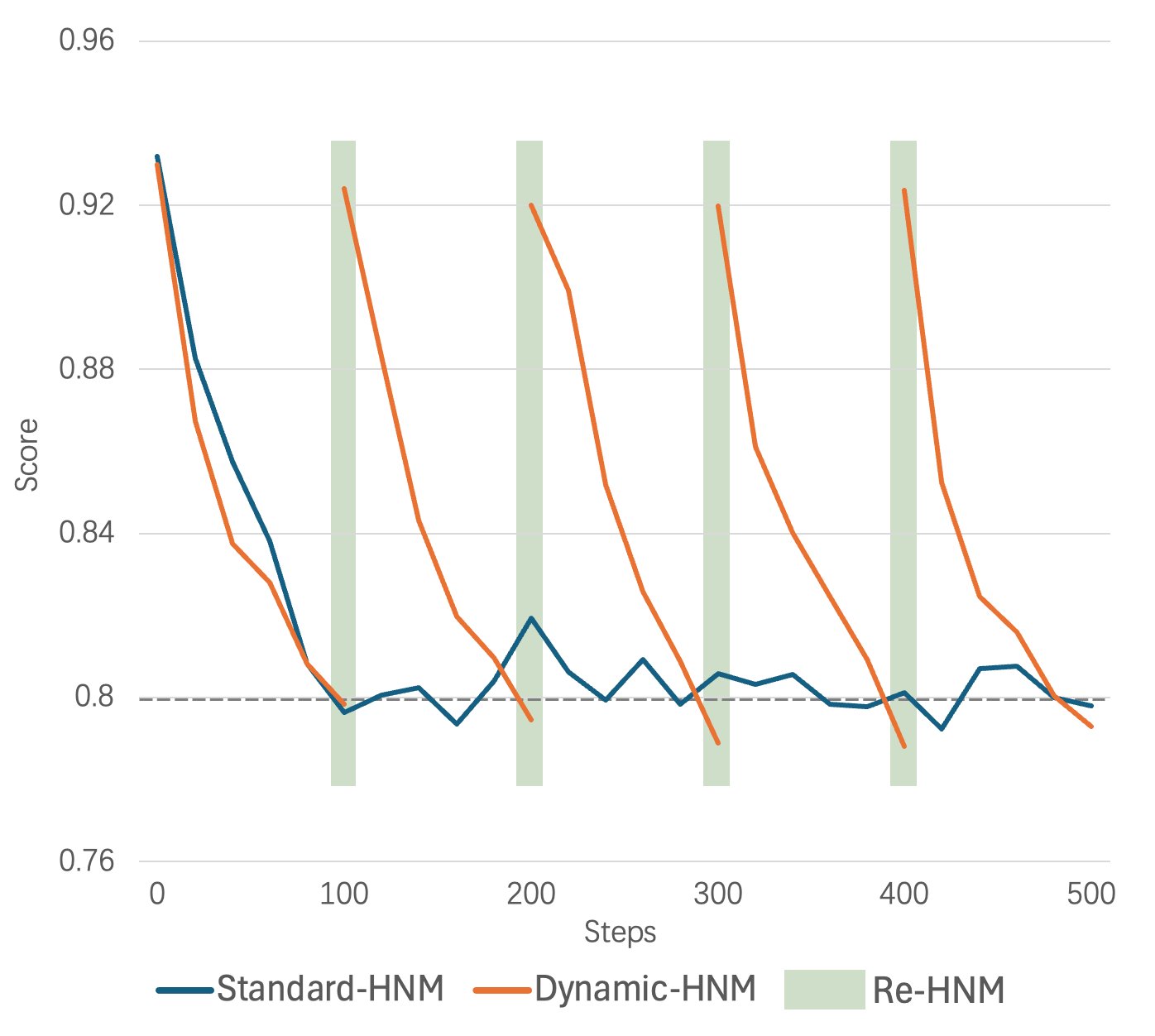} 
\caption{Dynamic Hard Negative Mining vs. Standard Hard Negative Mining: Score-Steps Curves. Hard negatives are checked every 100 steps. When the score multiplied by 1.15 is less than the initial score and the absolute value of the score is less than 0.8, we consider the negative example no longer difficult and replace it with a new hard negative.}
\end{wrapfigure}
\label{fig:DHNM}
Previous work has primarily focused on hard negative mining during the data preprocessing stage. For an embedding model with a given set of weights, the hard negatives are fixed. However, as training progresses and model weights are updated, the hard negatives corresponding to the current weights change. Hard negatives mined during the preprocessing stage may become less challenging after several training iterations.

Based on this insight, we propose a dynamic hard negative mining method. For each data point, we record the current average score of the hard negatives relative to the query. Every 100 iterations, if the score multiplied by 1.15 is less than the initial score and the absolute value of the score is less than 0.8, we consider the negative example no longer difficult and proceed with a new round of hard negative mining. During each dynamic hard negative mining, if hard negatives need to be replaced, we use  $(i-1) \times n + 10$ to $i \times n + 10$ cases as negative examples, where $i$ represents the i-th replacement and $n$ represents the number of hard negative cases used each time. The entire process incurs a cost equivalent to one step iteration.

Compared to In-Batch Negative InfoNCE Loss, we believe that higher quality hard negatives (more aligned with the current model weights) are more important. Figure.\ref{fig:DHNM} shows the score-step curves of positive and negative examples for dynamic hard negative mining versus standard hard negative mining. As seen, with increasing steps, the negative example scores in standard hard negative mining stop decreasing and start oscillating, indicating that the model has finished learning from that batch of negatives. In contrast, dynamic hard negative mining replaces the hard negatives once it detects that the negatives are no longer challenging for the model.


\subsection{Cross-GPU Batch Balance Loss}

To better leverage hard examples, we adopted the Cross-GPU Batch Balance Loss (CBB). Previous approaches~\cite{li2023challenging} typically assign a task to each batch randomly during the training process. 
For example, in iteration 0, samples from the STS task are selected, and the corresponding STS loss is used for backward to obtain gradients and update weights. In iteration 1, the Retri task might be assigned. We refer to this as sequential random task training. 
Such training often results in inconsistencies between the search space optimized in a single iteration and the global search space of the embedding model, which can cause oscillations during training.
These oscillations hinder the model's ability to converge to a global optimum, making it more challenging to achieve the best possible performance. We demonstrate this phenomenon in Sec.~\ref{CBB_loss_ana}.

\begin{wrapfigure}{h}{0.5\textwidth}
\vspace{-5pt}
\centering
\includegraphics[width=7cm]{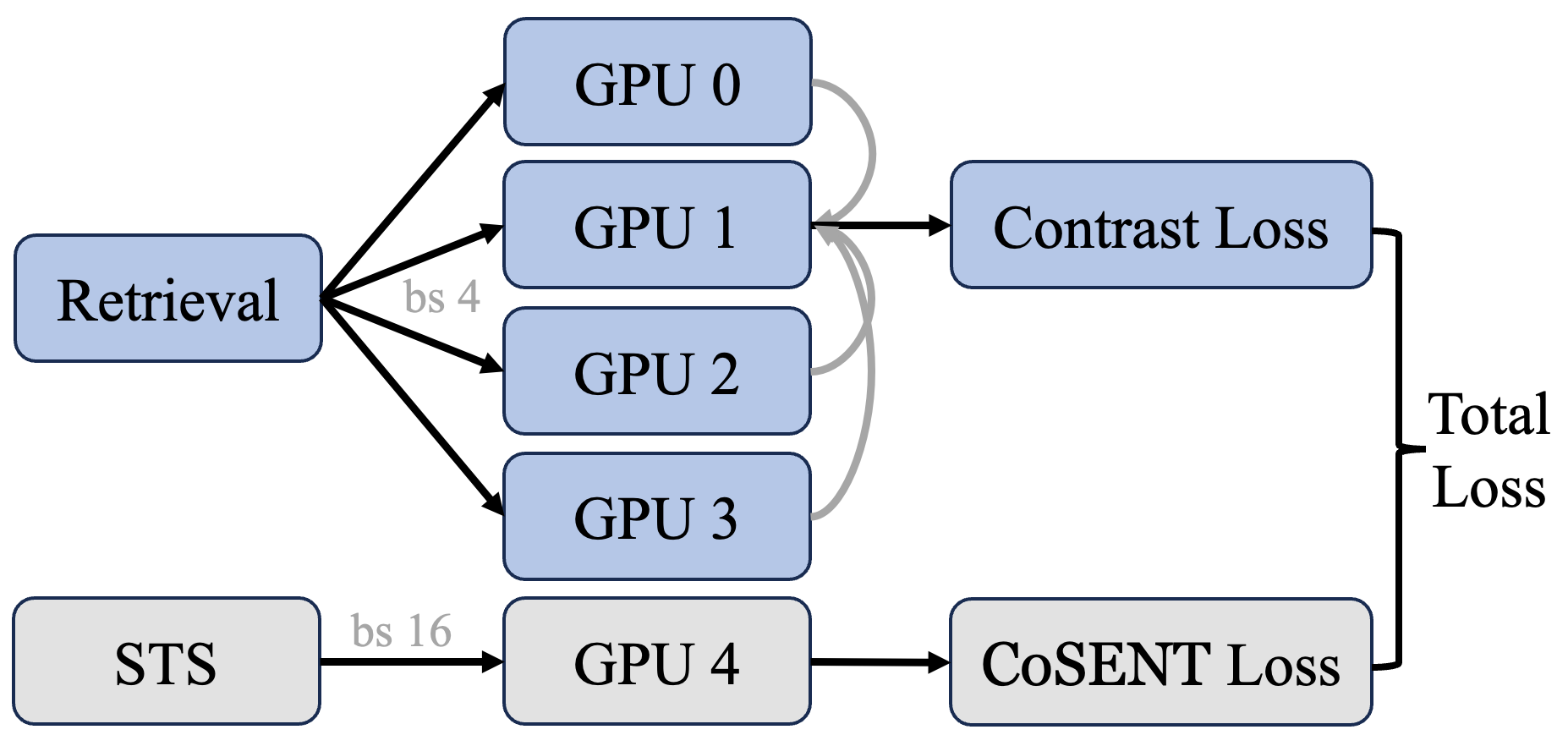} 
\caption{An example of cross-GPU batch balance Loss. For retrieval task, we leverage multiple GPUs to incorporate more negative examples. For STS task, we increase the batch size to include more cases for comparison.}
\label{fig:cross}
\end{wrapfigure}

To address this, we consider introducing each task in a balanced manner during each Forward-Loss-Backward-Update cycle to obtain a stable search space and minimize the discrepancy between the direction of a single model update and the global optimum. Therefore, the CBB strategy not only considers communication between different GPUs but also communication between different tasks, achieving better batch balancing. As shown in Figure.~\ref{fig:cross}, to utilize more hard examples in retrieval tasks, we ensure that each GPU (gpu0, gpu1, gpu2, gpu3) has different negative examples while sharing the same queries and positive examples. For the Retri task, each GPU calculates the loss for its corresponding batch, and the results are aggregated on gpu1. For the STS task, gpu4 runs the STS task and obtains the corresponding loss. Finally, the results are aggregated to compute the combined CBB Loss for the current iteration. The corresponding formula is as follows:

\begin{equation} \mathcal{L}_{\text{CBB}} = -\frac{1}{n} \sum_{i} \log \frac{\exp(s(x_i, y_i^{+})/\tau)}{\exp(s(x_i, y_i^{+})/\tau) + \sum_{k=1}^{N} \sum_{j=1}^{n} \exp(s(x_i, y_{j}^{-})/\tau)} +  \beta \times \mathcal{L}_{\text{cos}}
\end{equation}

where $s(x_i, y_i^{+})$ is a scoring function between query $x_i$ and positive text $y_i^{+}$, often defined as the cosine similarity, $N$ is the number of GPUs sharing the query $x_i$ and positive text $y_i^{+}$, and $\tau$ is the scale temperature. We set $\beta$ to 0.8 empirically.

\section{Experiments}
\subsection{Implementation details}
As most embedding models do, we also use the BERT large model~\cite{Jacob2018bert} as our base model and employ a linear layer to expand the dimensionality from 1024 to 1792. Consequently, the total number of parameters in the model is 326M. 
Additionally, inspired by OpenAI text-embedding-v3~\cite{text-embedding-v3}, we also utilize Matryoshka Representation Learning (MRL) techniques from~\cite{kusupati2022matryoshka} to achieve flexible dimension lengths. 
 
The model is trained with a maximum input length of 512 tokens. To enhance efficiency, mixed precision training and DeepSpeed ZERO-stage 1~\cite{rajbhandari2020zero} are utilized. 
For the pre-training stage, we use AdamW~\cite{loshchilov2017decoupled} optimizer and learning rate of 1e-5, with 0.05 warmup ratio and 0.001 weight decay. The batch size is set to 8. The entire pre-training process employs 64 Ascend 910B GPUs and 138 hours.
For the finetune stage, the MRL training representation dimensions are configured as 256, 512, 768, 1024, 1280, 1536 and 1792. The batch size is set to 4 for the retrieval task and 32 for the STS task. We used the same optimizer parameters and learning rate as in the pre-training phase. The entire fine-tuning process employs 16 Ascend 910B GPUs and takes 13 hours.
\begin{table}[h!]
    \centering
    \caption{Overview of the data sources used for pre-training.} 
    \begin{tabular}{l|ccc}
      \toprule
      Categories & Data Format & Prop & Numbers \\
      \midrule
      News & (title, content) & 27.3\% & 233M \\
      Knowledge Base & (question, answer) & 7.7\% & 66M \\
      Social Media & (title, content) & 39.9\% & 341M \\
      Web Page & (input, output) & 4.6\% & 39M \\
      Academic Paper & (title, content) & 6.0\% & 51M  \\
      Community QA & (question, answer) & 1.6\% & 14M \\
      Instruction datasets & (prompt, response) & 11.7\% & 100M \\
      LLM generated & (question, answer) & 1.2\% & 10M \\
      \bottomrule
    \end{tabular}
    \label{tab:pretrain_data}
\end{table}

\subsection{Datasets}

During the pre-training phase, we collected 0.75 billion text data pairs from the internet, categorized into title-content pairs, input-output pairs, and question-answer pairs. We also discovered that high-quality LLM instruction-tuning data, such as prompt-response pairs, can enhance the performance of embedding models after being filtered and screened according to rules. Additionally, we used LLMs to generate a batch of data utilizing existing text corpora. The detailed data description can be found in Table~\ref{tab:pretrain_data}.

\begin{table}[h!]
    \centering
    \caption{Data formats and quantities for different tasks.}
    \resizebox{\linewidth}{!}{
    \begin{tabular}{l|ccccccc}
      \toprule
      Tasks & Data Format & Loss & Numbers \\
      \midrule
      STS & (text, text pairs, score) & CoSENT Loss & 1.3M \\
      Retrieval & (text, text positive, text negative) & InfoNCE loss & 1.8M  \\
      STS generated & (text, text pairs, score) & CoSENT Loss & 0.6M \\
      Retrieval generated & (text, text positive, text negative) & InfoNCE loss & 0.5M\\
      \bottomrule
    \end{tabular}}
    \label{tab:finetune_data}
\end{table}
During the fine-tuning phase, to make the model more adaptable to various tasks, we selected common retrieval, classification, and STS (semantic textual similarity) datasets. For classification tasks, we merged them by considering data within the same category as text positive and data from different categories as text negative, and then consider them as retrieval tasks. The amount of data used in the fine-tuning phase is shown in the Table~\ref{tab:finetune_data}.

\subsection{CMTEB Results}
\begin{table}[h!]
    \centering
    \caption{Results on CMTEB. We report the average performance on six different tasks: Classification (CLS), Clustering (Cluster), Pair Classification (Pair CLS), Reranking (Rerank), Retrieval (Retri) and Semantic Textual Similarity (STS).}
    \resizebox{\linewidth}{!}{
    \begin{tabular}{l|ccccccc}
      \toprule
      Models & Average & CLS & Cluster & Rerank & Retri & STS & Pair CLS \\
      \midrule
      piccolo-large-zh-v2 & 70.95 & 74.59 & 62.17 & 70.00 & 74.36 & 63.50 & 90.24 \\
      IYun-large-zh & 71.04 & 74.18 & \textbf{66.35} & 69.30 & 73.56 & 63.23 & 90.87 \\
      zpoint-large-embedding-zh & 71.88 & 74.43 & 62.23 & 72.34 & 76.36 & 64.22 & 91.55 \\
      gte-Qwen2-7B-instruct & 72.05 & \textbf{75.09} & 66.06 & 68.92 & 76.03 & \textbf{65.33} & 87.48 \\
      xiaobu-embedding-v2 & 72.43 & 74.67 & 65.17 & 72.58 & 76.50 & 64.53 & 91.87 \\
      \midrule
      Conan-embedding & \textbf{72.62} & 75.03 & 66.33 & \textbf{72.76} & \textbf{76.67} & 64.18 & 91.66 \\
      \bottomrule
    \end{tabular}}
    \label{tab:cmteb}
\end{table}

MTEB (Massive Text Embedding Benchmark)~\cite{muennighoff2022mteb} is the most authoritative and popular benchmark for evaluating large-scale text embedding tasks. ~\cite{bge_embedding} created a Chinese embedding evaluation set known as CMTEB. CMTEB has 35 datasets spanning across 6 categories: Classification, Clustering, Pair Classification, Rerank, Retrieval, and STS. Table~\ref{tab:cmteb} presents a comparison of our models with others on the CMTEB benchmark. Our model surpasses the previous state-of-the-art model on nearly all tasks.

\subsection{Ablation Study}
\begin{table}[h!]
    \centering
    \caption{Ablation results on CMTEB. Baseline represents our result after the pre-training stage. Vanilla means directly finetune with standard InfoNCE loss and CoSENT loss. DHNM denotes using only the dynamic hard negative mining method. CBB Loss indicates using only the Cross-GPU Batch Balance Loss. }
    \resizebox{\linewidth}{!}{
    \begin{tabular}{c|ccccccc}
      \toprule
      Methods & Average & CLS & Cluster & Rerank & Retri & STS & Pair CLS \\
      \midrule
       Baseline & 62.9 & 60.4 & 62.7 & 70.4 & 63.2 & 55.2 & 87.3 \\
       Vanilla & 68.8 & 71.4 & 62.0 & 67.0 & 72.4 & 61.3 & 89.9  \\
       CBB Loss & 70.4 & 73.0 & 65.6 & 68.1 & 72.3 & 64.1 & 90.0 \\
       DHNM  & 71.2 & 74.4 & 66.2 & 69.0 & 73.8 & 63.5 & 90.4  \\
      \midrule
      Conan-embedding & \textbf{72.62} & \textbf{75.03} & \textbf{66.33} & \textbf{72.76} & \textbf{76.67} & \textbf{64.18} & \textbf{91.66} \\
      \bottomrule
    \end{tabular}}
    \label{tab:ablation}
\end{table}
To demonstrate the effectiveness of our method, we perform comprehensive ablation studies on the CMTEB benchmark. As shown in Table~\ref{tab:ablation}, it is clear that both dynamic hard negative mining and Cross-GPU Batch Balance Loss significantly outperform the vanilla method that directly fine-tunes the model. Notably, our conan-embedding model shows substantial improvements in retrieval and reranking tasks, indicating that the increased quantity and quality of negative examples allow the model to see more challenging negatives, thereby enhancing its recall capability.

\subsection{Analysis}\label{CBB_loss_ana}
\begin{wrapfigure}{h}{0.5\textwidth}
\vspace{-35pt}
\centering
\includegraphics[width=7cm]{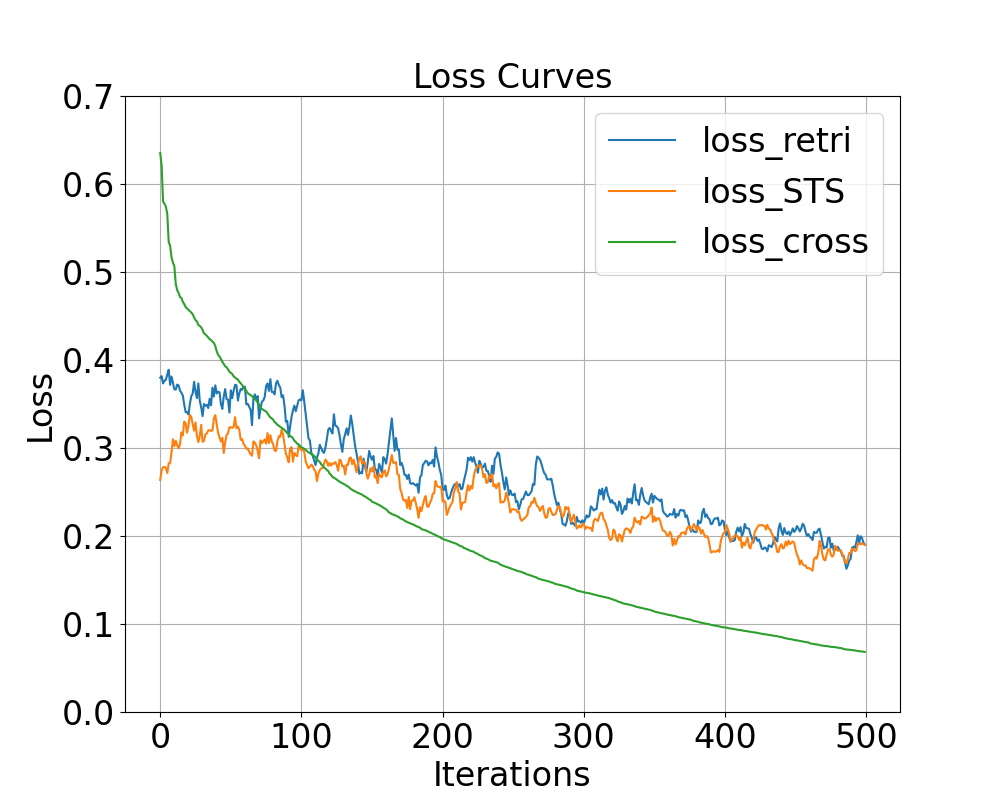} 
\caption{Comparison of Loss Curves Before and After Using the Cross-GPU Batch Balance Loss Method.}
\label{fig:loss}
\end{wrapfigure}

To better evaluate the effect of Cross-GPU Batch Balance Loss, we present the loss curves before and after using this loss in Figure~\ref{fig:loss}. The retri and STS loss represent the individual losses for the two tasks when trained together. It can be observed that the loss fluctuates significantly, decreases slowly, and does not decrease simultaneously. This indicates that there is a gap in the vector space between different tasks, and directly updating with different losses does not achieve optimal performance in optimization. The cross loss represents the use of Cross-GPU Batch Balance Loss. It can be seen that the loss is decreasing smoothly and continuously, and the final loss (0.08) is much smaller than the sum of the retri and STS losses (0.38).

\section{Conclusion}
In this paper, we introduced the conan-embedding model, designed to enhance the performance of embedding models by maximizing the quality and quantity of negative examples. Our approach revolves around two key innovations: dynamic hard negative mining and Cross-GPU balancing loss. The effectiveness of our approach is validated by our model's top-ranking performance on the Chinese leaderboard of the Massive Text Embedding Benchmark. We hope our method inspires more works to explore new ways of hard negative mining. The model has been uploaded to Huggingface: \href{https://huggingface.co/TencentBAC/Conan-embedding-v1}{Conan-embedding-v1}.
\newpage

\bibliography{colm2024_conference}
\bibliographystyle{colm2024_conference}


\end{document}